\documentclass[journal]{IEEEtran}

\usepackage[noadjust]{cite}
\usepackage{amssymb,comment,color}
\usepackage{amsmath}
\usepackage{amsthm}
\usepackage{graphicx}
\usepackage[caption=false, font=footnotesize]{subfig}
\usepackage[ruled,vlined]{algorithm2e}
\usepackage{multirow}

\newtheorem{proposition}{Proposition}

\ifCLASSINFOpdf
\else
\fi

\begin{document}

\title{OCD-FL: A Novel Communication-Efficient Peer Selection-based Decentralized Federated Learning}

\author{
    Nizar~Masmoudi and Wael~Jaafar,~\IEEEmembership{Senior~Member,~IEEE}
    
    \thanks{
        N. Masmoudi is with Université Paris Dauphine-PSL, Tunis Campus, Tunisia, e-mail: nizarmasmoudi@outlook.fr. W. Jaafar is with the department of Software and IT Engineering, École de technoogie supérieure (ÉTS) Montreal, QC, Canada, e-mail: wael.jaafar@etsmtl.ca.
    }
}

\markboth{Journal of \LaTeX\ Class Files,~Vol.~14, No.~8, XXX~2024}%
{Shell \MakeLowercase{\textit{et al.}}: Bare Demo of IEEEtran.cls for IEEE Journals}

\maketitle

\begin{abstract}
    The conjunction of edge intelligence and the ever-growing Internet-of-Things (IoT) network heralds a new era of collaborative machine learning, with federated learning (FL) emerging as the most prominent paradigm. With the growing interest in these learning schemes, researchers started addressing some of their most fundamental limitations. Indeed, conventional FL with a central aggregator presents a single point of failure and a network bottleneck. To bypass this issue, decentralized FL where nodes collaborate in a peer-to-peer network has been proposed. Despite the latter's efficiency, communication costs and data heterogeneity remain key challenges in decentralized FL. In this context, we propose a novel scheme, called opportunistic communication-efficient decentralized federated learning, a.k.a., OCD-FL, consisting of a systematic FL peer selection for collaboration, aiming to achieve maximum FL knowledge gain while reducing energy consumption. Experimental results demonstrate the capability of OCD-FL to achieve similar or better performances than the fully collaborative FL, while significantly reducing consumed energy by at least 30\% and up to 80\%.
\end{abstract}


\IEEEpeerreviewmaketitle

\vspace{-10pt}
\section{Introduction}
   With the increasing concerns around data privacy and continuous efforts to enhance the quality and speed of data processing, edge intelligence is becoming the new standard \cite{9596610}. An ever-growing Internet-of-Things (IoT) network is laying the groundwork for a massive edge environment that will revolutionize smart devices and networks. Thus, interest in collaborative machine learning (ML) has massively increased with Google's Federated Learning (FL) presented as one of the most promising paradigms \cite{9599369}. 
    Surveys \cite{9264412} focused on investigating the FL model while highlighting its key challenges, e.g., costly communication, resource heterogeneity, and data imbalance. Others attempted to tackle these problems. For instance, Yang \textit{et al.} proposed in \cite{9264742} a resource allocation model to minimize the energy consumption of clients under a latency constraint. Zhang \textit{et al.} presented in \cite{10000664} a relay-based topology where each client serves as a relay to assist distant clients in sharing their FL models. Their approach focused on maximizing each client's utility according to its serving role as a computational node and a relay. Furthermore, Wang \textit{et al.} focused on counterbalancing the bias introduced by non-IID data through a deep reinforcement learning algorithm that systematically selects clients to participate in each round  \cite{9155494}, while Han \textit{et al.} proposed an adaptive heterogeneity-aware scheduling to mitigate resource and data heterogeneity \cite{10020721}.

    As Beltrán \textit{et al.} highlighted in their survey \cite{10251949}, the aggregation server in a centralized Federated Learning (FL) setting serves as a single point of failure and a network bottleneck. \textcolor{blue}{Additionally, this centralized approach introduces security vulnerabilities, making it susceptible to network breaches and model corruption due to its status as a single point of attack. Furthermore, the aggregation server is tasked with combining the model parameters from all participating nodes, resulting in significant computational overhead. Finally, centralized FL may not be suitable for systems where components are dispersed with limited connectivity such as IoT networks, vehicular networks, and drone swarm networks.} These limitations pushed the proposal of a decentralized topology where clients communicate with each other inside a peer-to-peer network. Nevertheless, communication costs and data heterogeneity persist as key constraints in decentralized federated learning (D-FL). In this context, Zheng \textit{et al.} proposed an algorithm to balance between energy consumption and learning accuracy \cite{9498853}. Li \textit{et al.} focused on designing a robust solution in non-identically and distributed (non-IID) environments by achieving an effective clustered topology using client similarity and implementing a neighbor matching algorithm \cite{9954190}. Liu \textit{et al.} aimed to achieve a balance between communication efficiency and model consensus using multiple periodic local updates and inter-node communications \cite{9713700}. \textcolor{blue}{Du \textit{et al.} introduced in \cite{du2022gradientchannelawaredynamic} a dynamic device scheduling mechanism that optimizes the peer selection strategy and power allocation to improve the federated edge learning model accuracy. Their approach leverages the superposition characteristics of wireless channels to enhance model training at the server and proposes a method to measure local data importance based on the gradient of local model parameters, channel conditions, and energy consumption. Simulation results show that the proposed scheduling mechanism achieves high test accuracy, fast convergence rates, and robustness against different channel conditions.} \textcolor{blue}{In \cite{9801730}, Zhang \textit{et al.} proposed a blockchain and AI-based secure cloud-edge-end collaboration scheme coupled with a blockchain-empowered federated deep actor-critic-based task offloading algorithm to tackle the secure and low-latency computation offloading problem. Finally, Xiao \textit{et al.} addressed the time-varying dynamic network behavior by proposing a D-FL framework based on an inexact stochastic parallel random walk alternating direction method of multipliers, called ISPW-ADMM \cite{9475989}.} \\

    Despite the compelling results of D-FL, most published works
    establish their algorithms on a dense network of devices that do not consider mobility constraints. In a real-world setting, smart mobile devices, for instance, UAVs, tend to constitute a sparse graph where vertices are in continuous movement. This setting hinders model consensus across the entire network as each client performs a federated averaging procedure with a random fragment of the network in each FL round. Furthermore, the majority of the aforementioned papers rely on a bidirectional communication protocol between clients. This bidirectional exchange of knowledge can potentially improve the performances of one model at the expense of another, in addition to increasing communication costs.
    
    Motivated by the aforementioned observations, we propose an opportunistic communication-efficient decentralized federated learning (OCD-FL) scheme established on a sparse network of clients who follow different motion patterns. The main contributions are summarized as follows:
    \begin{enumerate}
        \item 
        Unlike previous works, we conduct our study for a sparse network of clients where each node can communicate only with its neighbors, i.e., nodes within its range of communication.
        Also, nodes' locations vary over time such that the neighbors of a given node change from one FL round to another.
        
        \item 
        We design a novel D-FL framework where each client makes a systematic decision to share its model with a neighbor aiming to enhance the latter's FL knowledge gain. Our approach is designed to get the maximum benefit from aggregation while saving as much energy as possible per FL client.
        
        \item 
        Using benchmark datasets under IID and non-IID scenarios, we implement our algorithm and baseline schemes, then run extensive simulations to demonstrate the efficiency of our solution compared to others, in terms of accuracy, loss, and energy. The obtained results demonstrate the high potential of OCD-FL.
    \end{enumerate}

    The paper is organized as follows. 
    Section \ref{sec:system-model} describes the system model. Section III describes D-FL. In
    section \ref{sec:design}, we expose our proposed method, while section \ref{sec:simulation-discussion} presents the simulation results. Finally, Section \ref{sec:conclusion} concludes the paper.

\vspace{-10pt}
\section{System Model} \label{sec:system-model}
    In this section, we present an overview of the adopted D-FL scheme by describing the network layout, the allocation strategy of data chunks across nodes, the local learning scheme, and the collaboration algorithm used in our design. Several notions are also introduced to help pave the path towards the proposed OCD-FL. Finally, a summary of the entire framework is described in Algorithm \ref{alg:d-fl}.

\vspace{-10pt}
    \subsection{Network Layout}
        We assume an ad hoc network of $N$ nodes. The network is represented using an undirected graph $\mathcal{G} = (\mathcal{N}, \mathcal{E})$ where $\mathcal{N} = \{1, 2, \dots, N\}$ is the set of nodes and $\mathcal{E}$ is the set of edges. Note that $N \triangleq |\mathcal{N}|$ denotes the number of nodes with $|\ .\ |$ being the cardinality of the set. Two nodes $(i, j) \in \mathcal{N}^2$ are connected if they are within each other's range of communication. This connection is denoted by $(i, j) \in \mathcal{E}$. We define by $\mathcal{K}_i$ the neighborhood of node $i$, i.e., the set of nodes within its range of communication, particularly $\mathcal{K}_i = \{j \in \mathcal{N}\ \text{s.t.}\ (i, j) \in \mathcal{E}\}$. We also define $K_i \triangleq |\mathcal{K}_i|$ the number of neighbors of node $i$. To emulate a real-world setting where nodes are mobile, the graph configuration changes at each FL round. Particularly, nodes change locations following several patterns, and therefore each one can connect to a different set of neighbors at each FL round. 

\vspace{-10pt}
    \subsection{Communication Model}
        We assume that each node is equipped with a single antenna used in half-duplex mode. Moreover, the random way-point mobility model is used to represent the change in clients' locations over time \cite{Johnson1996}. At any given FL round, the wireless channel between each pair of nodes $(i,k)$ is dominated by the Line-of-Sight (LoS) component, i.e., the channel path loss $G_{i,k}$ between node $i$ and a neighbor $k$ is written as (in dB)
        \begin{equation}       
        \label{eq:channel-gain-def}
            G_{i,k} = 10 \log_{10} \left( {P_{i,k}^r}/{P_i^t} \right), \; \forall i \in \mathcal{N}, \forall k \in \mathcal{K}_i, 
        \end{equation}
        where $P_{i,k}^r$ and $P_{i}^t$ are the received power at node $k$ and transmitted power by node $i$, respectively. Based on the Friis formula, the received power $P_{i,k}^r$ can be expressed by \cite{Friis}
        \begin{equation} 
              \label{eq:friis}
           P_{i,k}^r = P_i^t G_i^t G_k^r \left( {c}/{4 \pi f} \right)^2 {\left( d_{i,k} \right)^{-n}}, \; \forall i \in \mathcal{N}, \forall k \in \mathcal{K}_i
        \end{equation}
        where $G_i^t$ and $G_k^r$ are the antenna gains of the transmitter and receiver, respectively. $c$ denotes the speed of light, $f$ is the signal frequency, $d_{i,k}$ is the Euclidean distance between the transmitter and receiver, and $n$ is an environment variable.
        
        Using the Shannon-Hartley channel capacity formula, the achievable data rate can be given by (in bits/sec) \cite{Shannon}
        \begin{equation} 
              \label{eq:shannon}
            r_{i,k} = B_i\log_2\left( 1 + \frac{P_{i,k}^r} {N_0 B_i} \right), \; \forall i \in \mathcal{N}, \forall k \in \mathcal{K}_i
        \end{equation}
        where $B_i$ is the allocated bandwidth and $N_0$ is the power of the unitary additive white Gaussian noise (AWGN) in dBm/Hz.

\vspace{-11pt}
\section{Distributed FL: Background}

In this section, we describe the D-FL scheme, where each node may peer with its neighbors for FL aggregation.

\vspace{-11pt}
    \subsection{Dataset Distribution}
        Let $\mathcal{D}$ be the global dataset distributed across all nodes, i.e., each node $i \in \mathcal{N}$ owns a chunk of data denoted by $\mathcal{D}_i$ such that $\mathcal{D} = \mathcal{D}_1 \cup \mathcal{D}_2 \cup \dots \cup \mathcal{D}_N$ and $\mathcal{D}_i \cap \mathcal{D}_j = \emptyset \ \forall i \neq j$. We define $D \triangleq |\mathcal{D}|$ and $D_i \triangleq |\mathcal{D}_i|$, $\forall i \in \mathcal{N}$. The allocation of data chunks follows a Dirichlet distribution $\text{Dir}(\alpha)$ where $\alpha > 0$ is a parameter that determines the distribution and concentration of the Dirichlet. Dirichlet distributions are commonly used as prior distributions in Bayesian statistics and constitute an appropriate choice to simulate real-world data imbalance. It allows tuning distribution imbalance levels by varying $\alpha$ from low values (highly unbalanced) to high values (balanced). 

\vspace{-11pt}
    \subsection{Local Update}
        All nodes carry the same FL model architecture. We define, by $\mathbf{W}_i$ the model weight matrix of node $i$. To learn the intrinsic features of its local dataset, a node performs a local training operation. Assuming $\mathbf{X}_i$ as the input matrix of the learning model and $\mathbf{Y}_i$ is its target matrix, then the local optimization problem of node $i$ is defined as follows:
        \begin{equation}
              \label{eq:local-opt}
            \mathbf{W^*} = \underset{\mathbf{W}_i}{\arg\min}\ F(\mathbf{W}_i, \mathbf{X}_i, \mathbf{Y}_i),
        \end{equation}
        where $\mathbf{W^*}$ denotes the optimal (model weights) solution and $F$ denotes the loss function. The complexity of machine learning models and modern datasets translates into a complex shape of the loss function. With no guarantee of convexity, finding a closed-form solution to the problem \eqref{eq:local-opt} is usually intractable. As a result, gradient-based algorithms are used to solve it, namely Stochastic Gradient Descent (SGD) \cite{DBLP:journals/corr/Ruder16} and Adaptive Moment Estimation (Adam) \cite{kingma2017adam}.

\vspace{-10pt}
    \subsection{Federated Averaging}
        Collaboration among nodes is achieved through inter-node communications. Each node $i$ transmits its FL model to a set of neighbors, called peers, and denoted $\mathcal{R}_i$, such that $\mathcal{R}_i \subseteq \mathcal{K}_i$. Upon reception, each node carries out a federated averaging operation. Specifically, assuming $R_i \triangleq |\mathcal{R}_i|$ is the number of peers, federated averaging is performed as follows:
        \begin{equation}
        \label{eq:FLavg}
            \mathbf{W}_i^{\rm {agg}} = \frac{1}{R_i + 1} \left(\sum_{j \in \mathcal{R}_i}\mathbf{W}_j + \mathbf{W}_i\right).
        \end{equation}
        \textcolor{blue}{OCD-FL is built on an asynchronous wireless network 
        that leverages orthogonal frequency-division multiple access (OFDMA) to avoid the interference of concurrent transmissions \cite{Shah2021}. This allows clients to send their updates as soon as they are ready without waiting for stragglers \cite{Li_2020}. Nevertheless, the local aggregation has to wait for the end of the training round before aggregating the local model with the received ones.} In Algo. 1, we summarize the proposed D-FL.

    \begin{algorithm}[t]
        \caption{D-FL scheme.}
        \label{alg:d-fl}
        \SetKwInOut{Input}{Input}
        \SetKwInOut{Output}{Output}
        \SetKw{KwBy}{by}
        \Input{
            Graph $\mathcal{G = (\mathcal{N, \mathcal{E}})}$, dataset $\mathcal{D}$, Dirichlet parameter $\alpha$, number of rounds $Q$.
        }
        \For{$i \in \mathcal{N}$}{
            Allocate data chunk $\mathcal{D}_i$ to node $i$ following $\text{Dir}(\alpha)$\;
            Initialize model weight $\mathbf{W}_i$ of node $i$\;
        }
        \For{$r \gets 1$ \KwTo $Q$}{
            \For{$i \in \mathcal{N}$}{
                Move node $i$ to a different location\;
                Simultaneously perform local training (satisfy (\ref{eq:local-opt})) and receive models from other nodes\;
                Update $\mathcal{K}_i$ and select set of peers $\mathcal{R}_i \subseteq \mathcal{K}_i$\;
                Transmit local model to peers\;
                Execute federated averaging using (\ref{eq:FLavg})\;
            }
        }
        Return $\textbf{W}_i$, $\forall i \in \mathcal{N}$
    \end{algorithm}

\section{Proposed OCD-FL Scheme} \label{sec:design}

    The proposed OCD-FL is based on Algo. 1. However, it designs a specific peer selection mechanism that maximizes the benefit of peer-to-peer aggregation, while saving communication energy. To formulate the peer selection problem, we preliminarily define the energy consumption and knowledge gain expressions, needed for the objective design.     

\vspace{-10pt}
\subsection{Energy Consumption}
        Assuming that $S$ is the size of data a node transmits to peers, the transmission energy can be expressed by (Joules)
        \begin{equation}
        \label{eq:energy-final}
            E_{i,k}= \frac{ P_i^t \; S}{r_{i,k}}= \frac{P_i^t \; S}{B_i \log_2 \left(1+ \frac{P_{i,k}^r}{N_0 B_i} \right) }, \; \forall i \in \mathcal{N}, \forall k \in \mathcal{K}_i 
        \end{equation}
        Energy is positive and increases with distance. Thus, assuming that the communication range is $d_i^{\max}$\textcolor{blue}{\footnote{\textcolor{blue}{The communication range ensures that a transmission from node $i$ to node $j$ occurs only if the distance between them $d_{i,j}\leq d_i^{\max}$}.}}, the energy consumed by node $i$ with a node located at its range edge, $E_i^{\max}$, is 
        \begin{equation}
            E_i^{\max} = {P_i^t \; S} / \left( {B_i \log_2\left( 1 + \frac{ P_{i,\max}^r}{N_0 B_i} \right)} \right),\; \forall i \in \mathcal{N},
        \end{equation}
        where $P_{i,\max}^r= P_i^t G_i^t G^r \left( \frac{c}{4 \pi f} \right)^2 \left(d_i^{\max} \right)^{-n}$, and $G^r$ is the antenna gain of the neighbor located at distance $d_i^{\max}$ from node $i$. Accordingly, energy can be scaled with min-max normalization as $\Tilde{E}_{i,k} = {E_{i,k}}/{E_i^{\max}}$, $\forall i \in \mathcal{N}$, i.e., $\Tilde{E}_{i,k} \in [0,1]$.

\vspace{-10pt}    
    \subsection{Knowledge Gain}
        Although federated averaging remains an efficient FL collaboration method, low-performing models may negatively influence their peers and thus degrade the results of high-performing models. The latter, however, offer a good opportunity for low-performing models to progress further and improve their efficiency. This parasitic exchange between models may hinder model consensus. The following propositions highlight this phenomenon:
        \begin{proposition} \label{theorem:knowledge-gain}
            Let $\mathbf{W^*}$ be the optimal solution of problem \eqref{eq:local-opt}, while $\mathbf{W_1}$ and $\mathbf{W_2}$ are the weight matrices of two different models. For convenience, model efficiency is assumed analogous to its similarity with the optimal solution. Also, the model defined by $\mathbf{W_1}$ outperforms the one of $\mathbf{W_2}$. Hence, 
            \begin{equation} 
             \label{eq:sup-assumption}
                \Vert \mathbf{W}^* - \mathbf{W}_1 \Vert \le \Vert \mathbf{W}^* - \mathbf{W}_2 \Vert.
            \end{equation}
Now, we can deduce the following statements:
            
             \begin{subequations}
             \small 
                \begin{equation} \label{eq:kg-1}
                    \Vert \mathbf{W}^* - \mathbf{W}^{\rm agg} \Vert \le \Vert\mathbf{W}^* -\mathbf{W}_2 \Vert,
                \end{equation}
                \begin{equation} \label{eq:kg-2}
                    \Vert \mathbf{W}^* -\mathbf{W}_1 \Vert - \frac{1}{2} \Vert \mathbf{W}_2 - \mathbf{W}_1 \Vert \le \Vert \mathbf{W}^* - \mathbf{W}^{\rm agg} \Vert,
                \end{equation}
                \begin{equation} \label{eq:kg-3}
                    \Vert \mathbf{W}^* - \mathbf{W}^{\rm agg} \Vert \le \Vert \mathbf{W}^* - \mathbf{W}_1 \Vert + \frac{1}{2} \Vert \mathbf{W}_2 - \mathbf{W}_1 \Vert,
                \end{equation}
            \end{subequations}
            where $\mathbf{W}^{\rm agg}={\left(\mathbf{W}_1+\mathbf{W}_2\right)}/{2}$.
        \end{proposition}

        \begin{IEEEproof}
            Using Cauchy-Schwarz inequality
            and \eqref{eq:sup-assumption}, the proof of \eqref{eq:kg-1} is as follows:
            \begin{equation}
                \begin{split}
                    & \Vert \mathbf{W}^* - \mathbf{W}^{\rm agg} \Vert = \Vert \mathbf{W}^* - \frac{\mathbf{W}_1 +\mathbf{W}_2}{2} \Vert = \frac{1}{2} \Vert 2 \mathbf{W}^* - \mathbf{W}_1 - \mathbf{W}_2 \Vert \\
                     &\le \frac{1}{2} \left(\Vert \mathbf{W}^* - \mathbf{W}_1 \Vert + \Vert \mathbf{W}^* - \mathbf{W}_2 \Vert \right) \\
                     &\le \frac{1}{2} \left(\Vert \mathbf{W}^* - \mathbf{W}_2 \Vert + \Vert \mathbf{W}^* - \mathbf{W}_2 \Vert \right) \le \Vert \mathbf{W}^* - \mathbf{W}_2 \Vert. \\
                \end{split}
            \end{equation}

            Since \eqref{eq:kg-2} and \eqref{eq:kg-3} derive from the reverse Cauchy-Schwarz inequality, 
            their joint proof is as follows: 
            \begin{equation}
            \label{eq:ineq}
                \begin{split}
                    & \Bigl| \Vert \mathbf{W}^* - \mathbf{W}^{\rm agg} \Vert - \Vert \mathbf{W}^* - \mathbf{W}_1 \Vert\Bigr| \le \Vert \mathbf{W}^{\rm agg} - \mathbf{W}_1 \Vert \\
                     &\le \Vert \frac{\mathbf{W}_1 +\mathbf{W}_2}{2} - \mathbf{W}_1 \Vert \le \frac{1}{2} \Vert \mathbf{W}_2 - \mathbf{W}_1 \Vert.
                \end{split}
            \end{equation}

            By extending \eqref{eq:ineq}, we obtain,
            \begin{equation}
            \begin{split}
               & - \frac{1}{2} \Vert \mathbf{W}_2 - \mathbf{W}_1 \Vert \le \Vert \mathbf{W}^* - \mathbf{W}^{\rm agg} \Vert - \Vert \mathbf{W}^* - \mathbf{W}_1 \Vert \\
            \Leftrightarrow 
            & - \frac{1}{2} \Vert \mathbf{W}_2 - \mathbf{W}_1 \Vert + \Vert \mathbf{W}^* - \mathbf{W}_1 \Vert \le \Vert \mathbf{W}^* - \mathbf{W}_{\rm agg} \Vert, \\
              & \text{and} \\
           & \frac{1}{2} \Vert \mathbf{W}_2 - \mathbf{W}_1 \Vert \geq \Vert \mathbf{W}^* - \mathbf{W}^{\rm agg} \Vert - \Vert \mathbf{W}^* - \mathbf{W}_1 \Vert
                 \\
              \Leftrightarrow &    
                \frac{1}{2} \Vert \mathbf{W}_2 - \mathbf{W}_1 \Vert + \Vert \mathbf{W}^* - \mathbf{W}_1 \Vert \geq \Vert \mathbf{W}^* - \mathbf{W}^{\rm agg} \Vert. \nonumber 
                \end{split}
        \end{equation}
            
            Thus, statements \eqref{eq:kg-2} and \eqref{eq:kg-3} are obtained.
        \end{IEEEproof}

        \textit{Proposition} \ref{theorem:knowledge-gain} confirms that low-performing models always benefit from high-performing models, while the opposite is not always true. Indeed, a low-performing model may hinder a high-performing one especially when models' dissimilarity is significant. Accordingly, we introduce a \textit{knowledge gain} measure to identify peers with low-performing and high-performing models. The knowledge gained by $k$ when receiving the model of node $i$ is defined as  
        \begin{equation} \label{eq:knowledge-gain}
            \gamma_{i,k} = \max(l_k - l_i,\ 0),\; \forall i \in \mathcal{N}, \; \forall k \in \mathcal{K}_i,
        \end{equation}
        where $l_k$ and $l_i$ are the loss measures of $k$ and $i$, respectively. $(l_k - l_i)$ is an underlying component that measures the \textcolor{blue}{model} performance disparity between $i$ and $k$. $l_k < l_i$ indicates that neighbor $k$'s performance outperforms that of node $i$, thus $\gamma_{i,k}=0$, and no benefit is gained from peering\textcolor{blue}{\footnote{\textcolor{blue}{While loss and accuracy are important metrics to evaluate local model training, knowledge gain, defined with the loss metric, is used with energy consumption as criteria to dictate the peer selection and aggregation strategy.}}}. 

        Since $\gamma_{i,k}$ is computed using the loss functions, its values are unbounded. To fit within our objective, we propose an exponential normalization such that the scaled knowledge gain is $\Tilde{\gamma}_{i,k} = \Gamma \left(\max \left( l_k-l_i,0 \right) \right)= 1 - \exp \left(-\mu \cdot \max \left(l_k - l_i,\ 0 \right) \right)$,
        where $\mu > 0$ determines the slope of exponential scaling. 

\vspace{-10pt}    
    \subsection{Problem Formulation}
    
\begin{figure}[t]
            \centering
            \includegraphics[width=0.7\columnwidth]{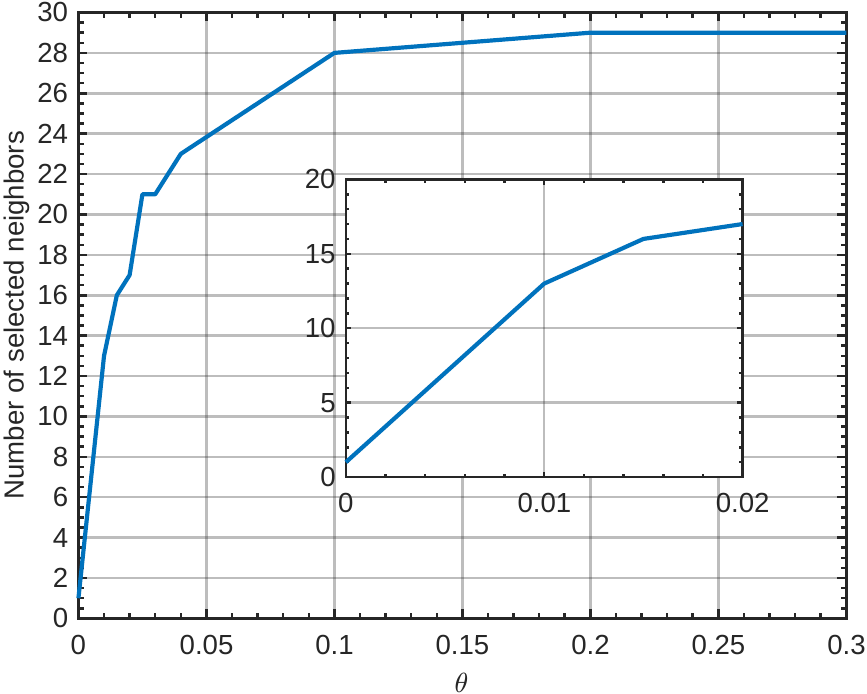}
            \caption{Effect of regularization on neighbor selection rate.}
            \label{fig:regularization}
        \end{figure}

        We formulate our problem as a node-specific multi-objective optimization problem. The goal is to efficiently select neighbors for collaboration taking into account the amount of energy required for transmission as well as the knowledge gained by neighbors as a result of the collaboration. Hence, for a given node $i$, we state the related problem as follows:
        \begin{subequations} 
        \label{eq:main-opt}
            \begin{align}
                \underset{w_k}{\max} \quad & \frac{
                    \sum_{k \in \mathcal{K}_i} \sigma(w_k) \Tilde{\gamma}_{i,k}
                }{
                    \sum_{k \in \mathcal{K}_i} \sigma(w_k) \Tilde{E}_{i,k}
                } + \theta \Vert w_k \Vert_2 \tag{\ref{eq:main-opt}} \\
                \text{s.t.} \quad & 1 \le \sum_{k \in \mathcal{K}_i} \sigma(w_k) \le K_i \label{eq:opt-const-a}
            \end{align}
        \end{subequations}       
        \noindent
        where $(w_k)_{k \in \mathcal{K}_i}$ are the selection model's trainable parameters, $\sigma(\cdot)$ denotes the sigmoid function, $\theta$ is a regularization parameter and $|| \cdot ||_2$ is the Euclidean norm. \textcolor{blue}{In contrast to L2-regularization in ML that aims to minimize the number of parameters in a system \cite{cortes2012l2regularizationlearningkernels}, we use $\theta ||w_k||_2$ with the maximization function to promote the selection of a higher number of D-FL neighbors by increasing the magnitude of the $k^{th}$ model's parameters $(w_k)_{k \in \mathcal{K}_i}$. $\theta$ is a parameter that controls the strength of the regularization effect. This component is necessary as empirical studies have shown that without regularization, the problem is reduced to selecting a single neighbor, enough to avoid an indeterminate form of the objective function, while obtaining a high knowledge gain-to-energy ratio. This hinders the local model’s ability to generalize. 
        In Fig. \ref{fig:regularization}, we plot the number of selected neighbors as a function of $\theta$ for a specific D-FL node with 30 neighbors, generated using random values for knowledge gain and energy consumption. Results demonstrate the importance of the regularization term to avoid selecting a single neighbor.}
        Constraint \eqref{eq:opt-const-a} guarantees that at least one neighbor is selected to avoid an indeterminate form while asserting that at most all neighbors are selected. For the sake of simplicity, we define by $\beta_k = \sigma(w_k)$ the probability that neighbor $k$ is selected as a peer. The objective is to learn $(w_k)_{k \in \mathcal{K}_i}$ for each node $i$ and, according to its $(\beta_k)_{k \in \mathcal{K}_i}$, decides under a certainty threshold the neighbors that will be peered. Since the objective function of \eqref{eq:main-opt} is not concave, it cannot be solved directly. 
        Nevertheless, \eqref{eq:main-opt} is differentiable, thus rendering its resolution with gradient-based algorithms feasible.

\section{Simulation Experiments and Results} \label{sec:simulation-discussion}


    \subsection{Simulation Setup}
        OCD-FL is implemented using Torch inside a Python environment. We adopt a sparse topology,
        where $N=20$ nodes are randomly placed on a 2-dimensional bounded rectangular surface. For each node $i$, $P_i^t$ and $B_i$ are uniformly distributed in the intervals $[10, 21]$ dBm and $[5, 20]$ MHz, respectively, $\forall i \in \mathcal{N}$. Antenna gains are $G_i^t=G_i^r=0$ dBi, $\forall i \in \mathcal{N}$. The signal frequency $f=1$ GHz, $c=3 \cdot 10^8$ m/sec, $d_i^{\max} = 2$ km, $\forall i \in \mathcal{N}$, {the size of data is $S=87$ Kbits (MNIST) and $S=23$ Mbits (CIFAR-10), $n=2$ (suburban), and $\mu=2$}. 

        \begin{figure}[t]
            \centering
            \subfloat[Accuracy (IID scenario)]{\includegraphics[width=0.5\linewidth]{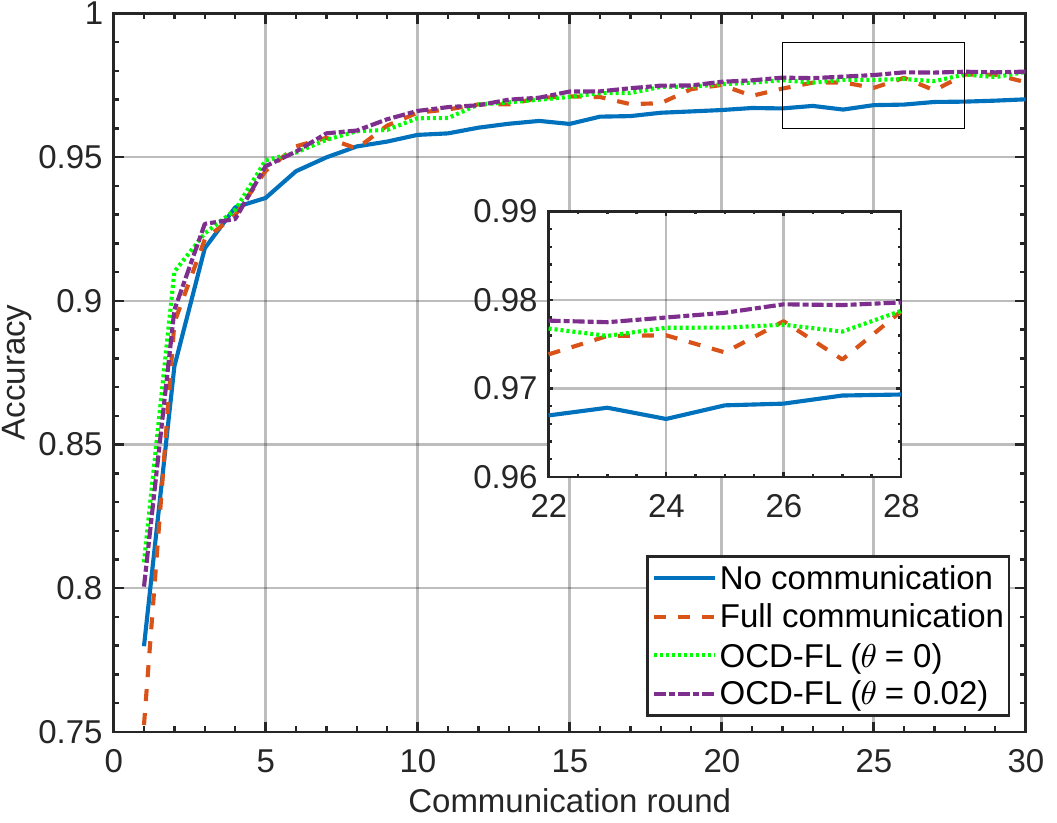}\label{fig:mnist-iid-acc}}%
            \hfil
            \subfloat[Loss (IID scenario)]{\includegraphics[width=0.5\linewidth]{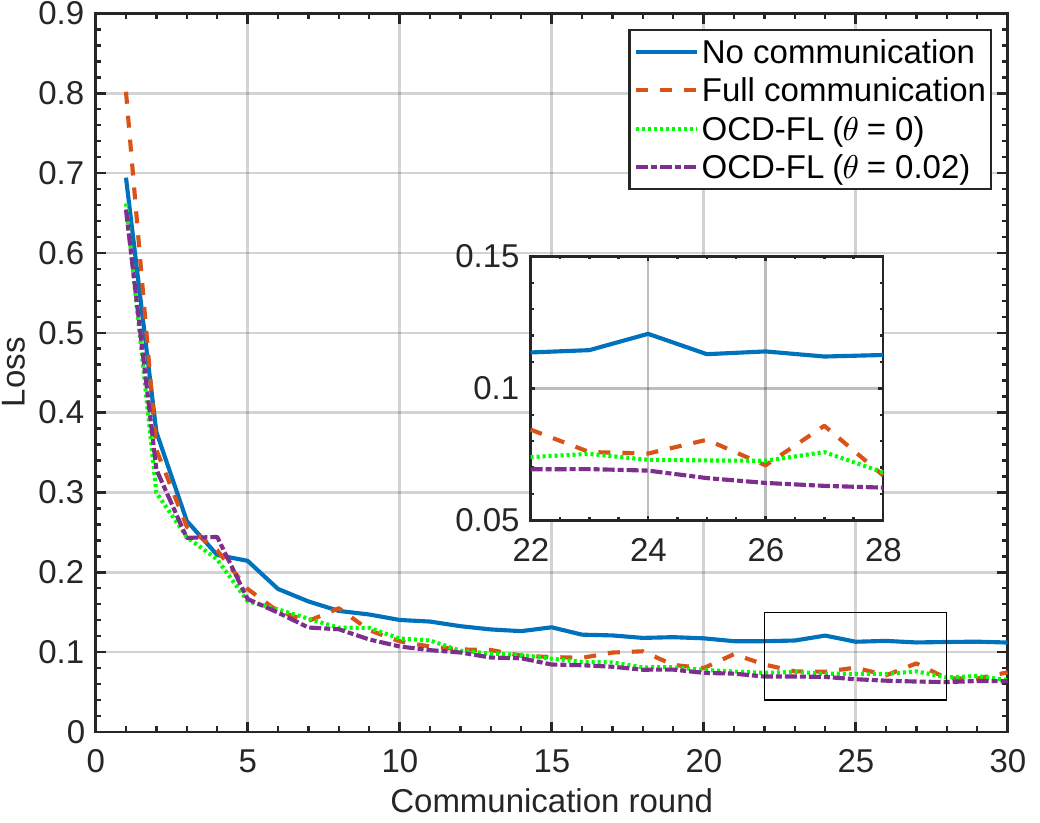}\label{fig:mnist-iid-loss}}%
            \vskip\baselineskip
            \subfloat[Accuracy (non-IID scenario)]{\includegraphics[width=0.5\linewidth]{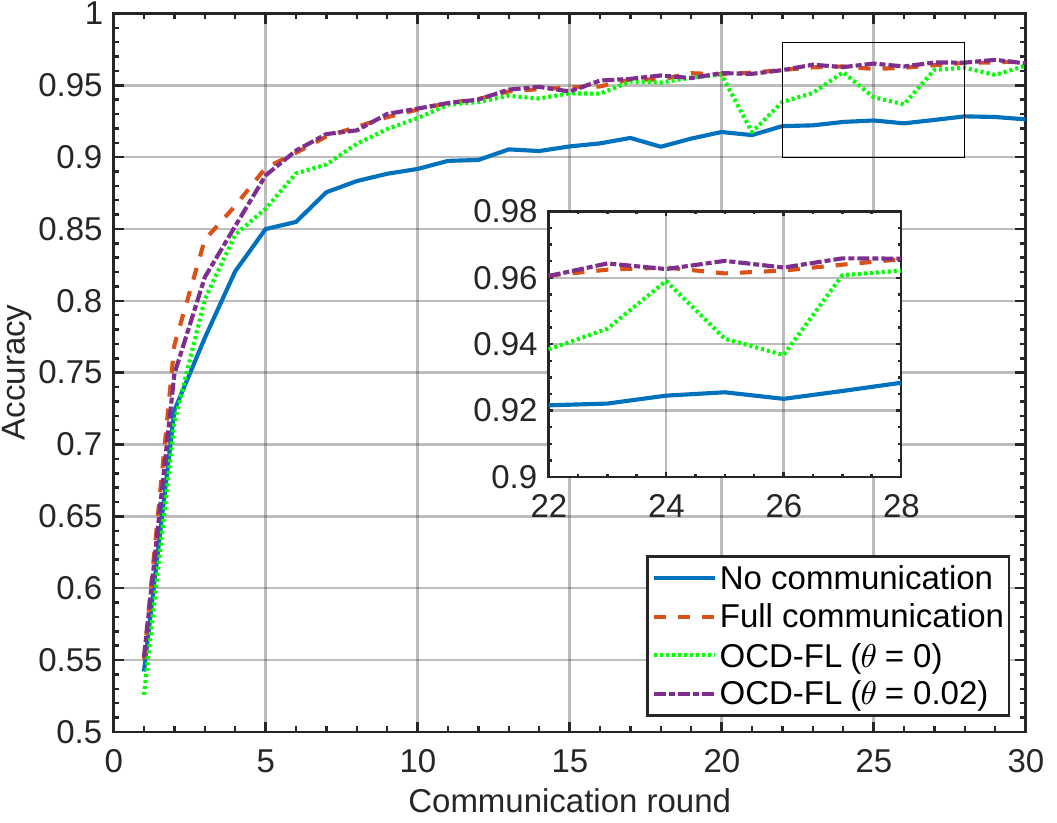}\label{fig:mnist-niid-acc}}%
            \hfil
            \subfloat[Loss (non-IID scenario)]{\includegraphics[width=0.5\linewidth]{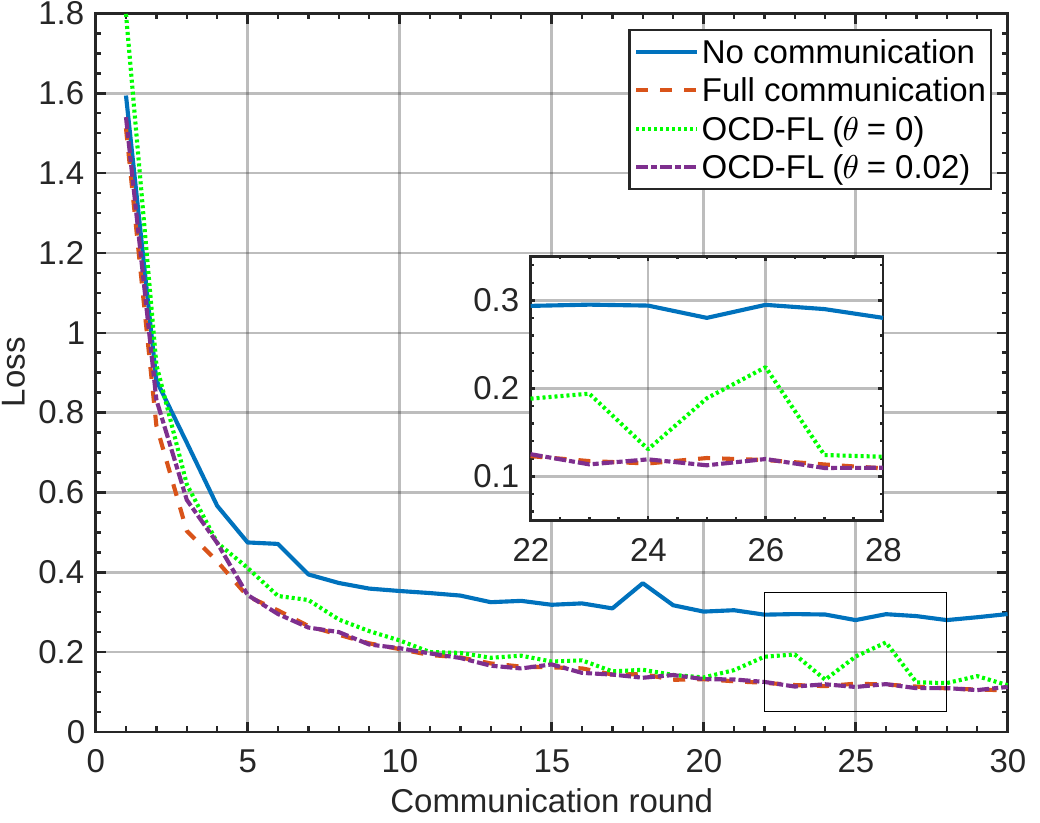}\label{fig:mnist-niid-loss}}%
            
            \caption{Avg. accuracy and loss (MNIST, different schemes).}
            \label{fig:mnist-eval}
        \end{figure}
        
        Our experiments are performed on two different datasets, MNIST and CIFAR-10 \cite{6296535,article}. Although both datasets consist of 60,000 training samples and 10,000 test samples, MNIST has square images ($28 \times 28 \times 1$ pixels) of handwritten digits (from 0 to 9), while CIFAR-10 contains colored square images ($32 \times 32 \times 3$ pixels) of 10 different object classes. Training \textcolor{blue}{dataset is split into equally-sized $N=20$ subsets following Dirichlet distribution with $\alpha = 1$ (non-IID scenario, i.e., number of samples in classes are significantly unbalanced), and $\alpha = 100$ (IID scenario, i.e., number of samples per class are approximately equal). In any case, the testing subset is IID to avoid non-biased evaluations.\footnote{\textcolor{blue}{With a non-IID testing dataset, overfitting is a prominent risk that yields biased evaluations of local models, thus hindering the performance of D-FL.}}}
         
        
        Different models are implemented to fit adequately each dataset. For MNIST, we used a model with 2 $5 \times 5$ convolutional neural network (CNN) hidden layers and ReLU activation functions, for a total number of parameters of 21,840. For CIFAR-10, the model is more complex with six $3 \times 3$ CNN layers, for a total number of parameters of 5,852,234. 
        During local updates, an NVIDIA Tesla T4 processing unit, along with a CUDA environment, was used to speed up computations.

 \begin{figure}[t]
            \centering
            \subfloat[Accuracy (IID scenario)]{\includegraphics[width=0.5\linewidth]{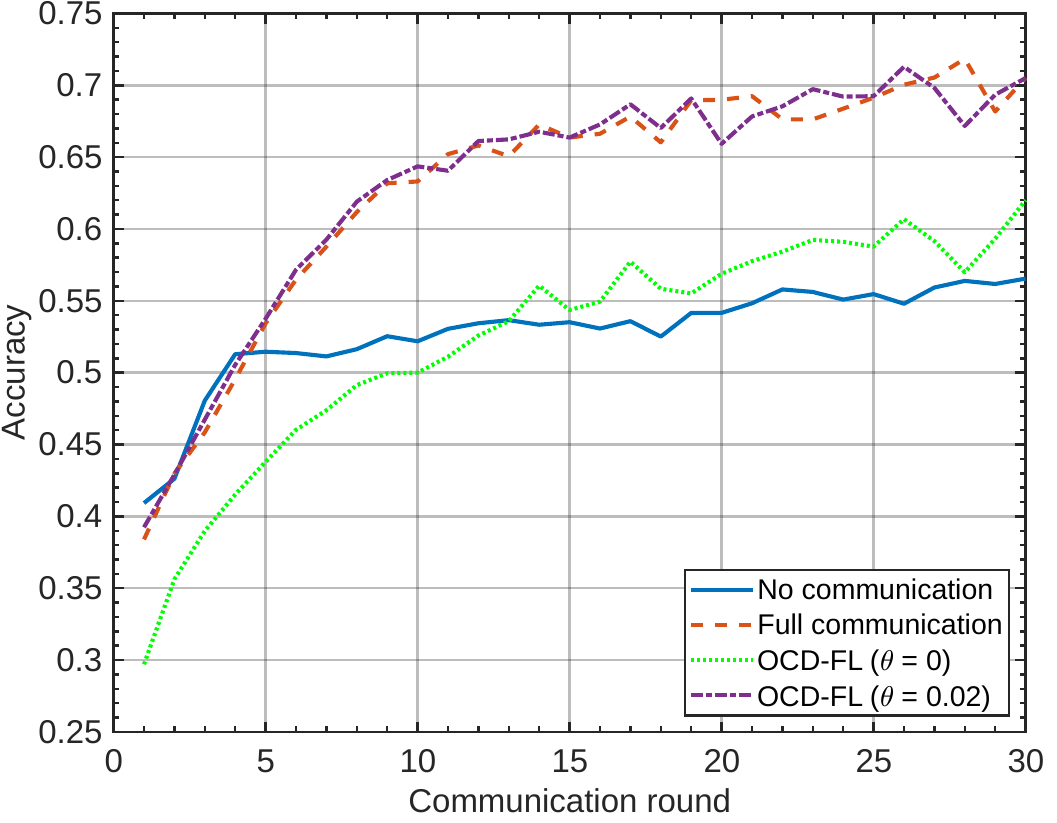}\label{fig:cifar-iid-acc}}%
            \hfil
            \subfloat[Loss (IID scenario)]{\includegraphics[width=0.5\linewidth]{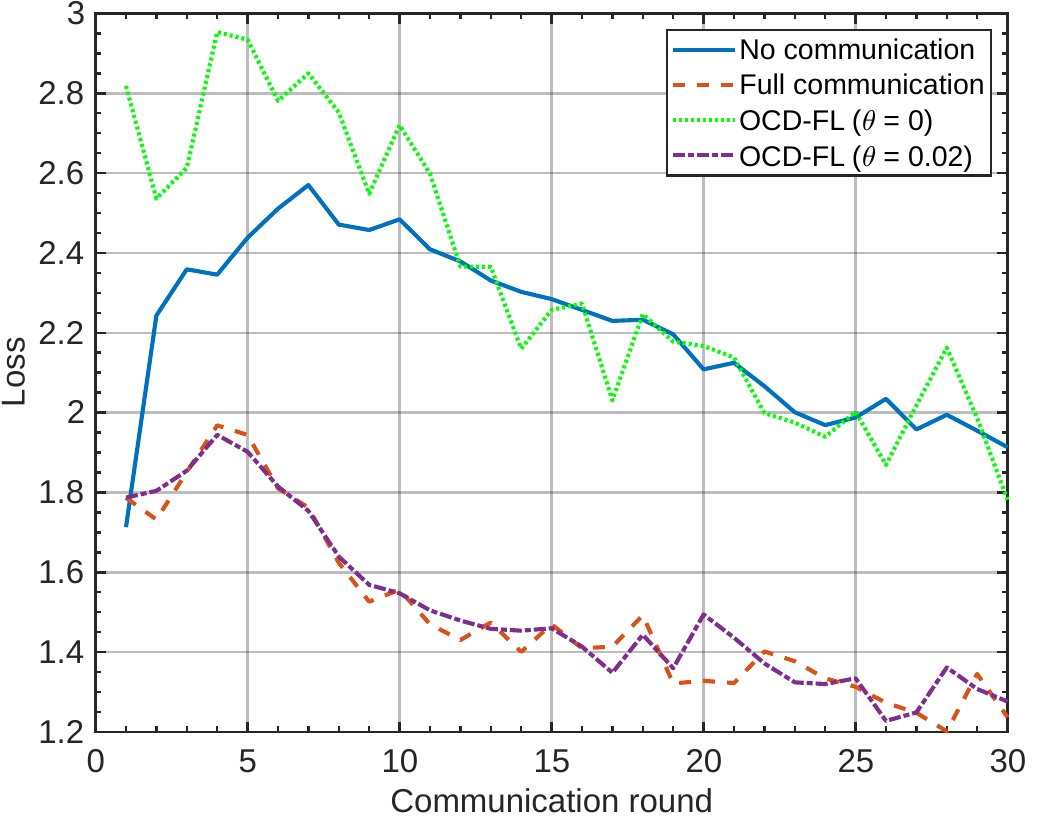}\label{fig:cifar-iid-loss}}%
            \vskip\baselineskip
            \subfloat[Accuracy (non-IID scenario)]{\includegraphics[width=0.5\linewidth]{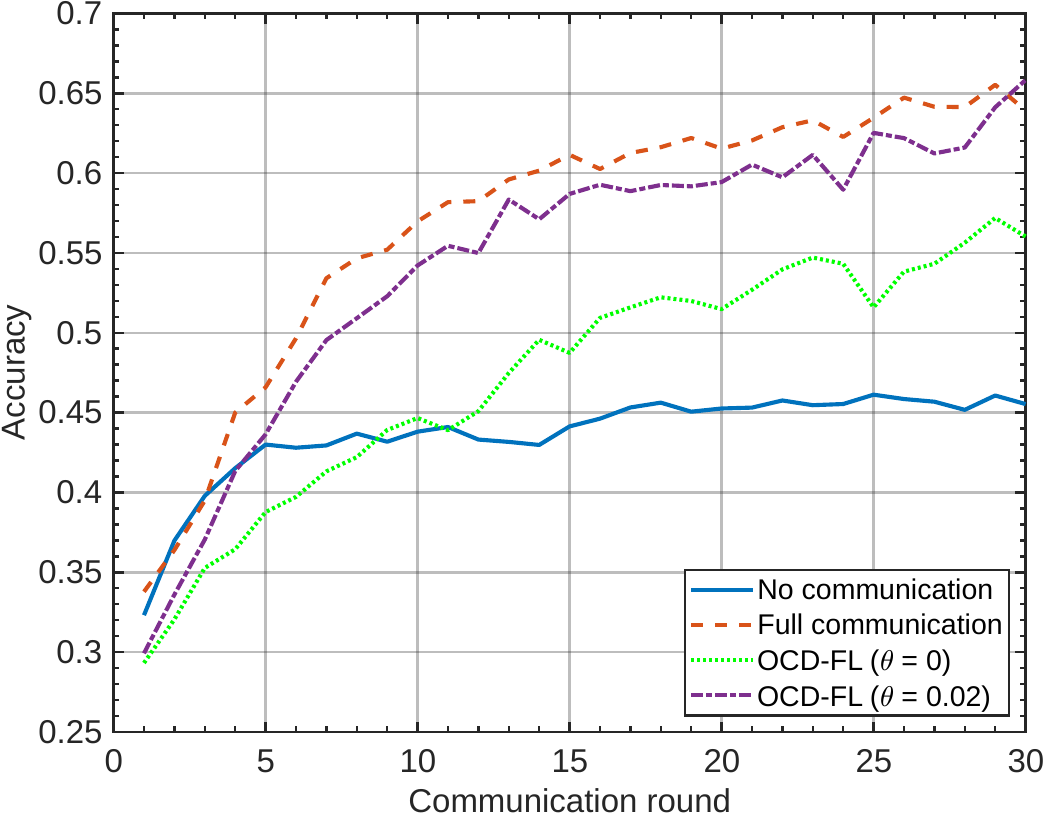}\label{fig:cifar-niid-acc}}%
            \hfil
            \subfloat[Loss (non-IID scenario)]{\includegraphics[width=0.5\linewidth]{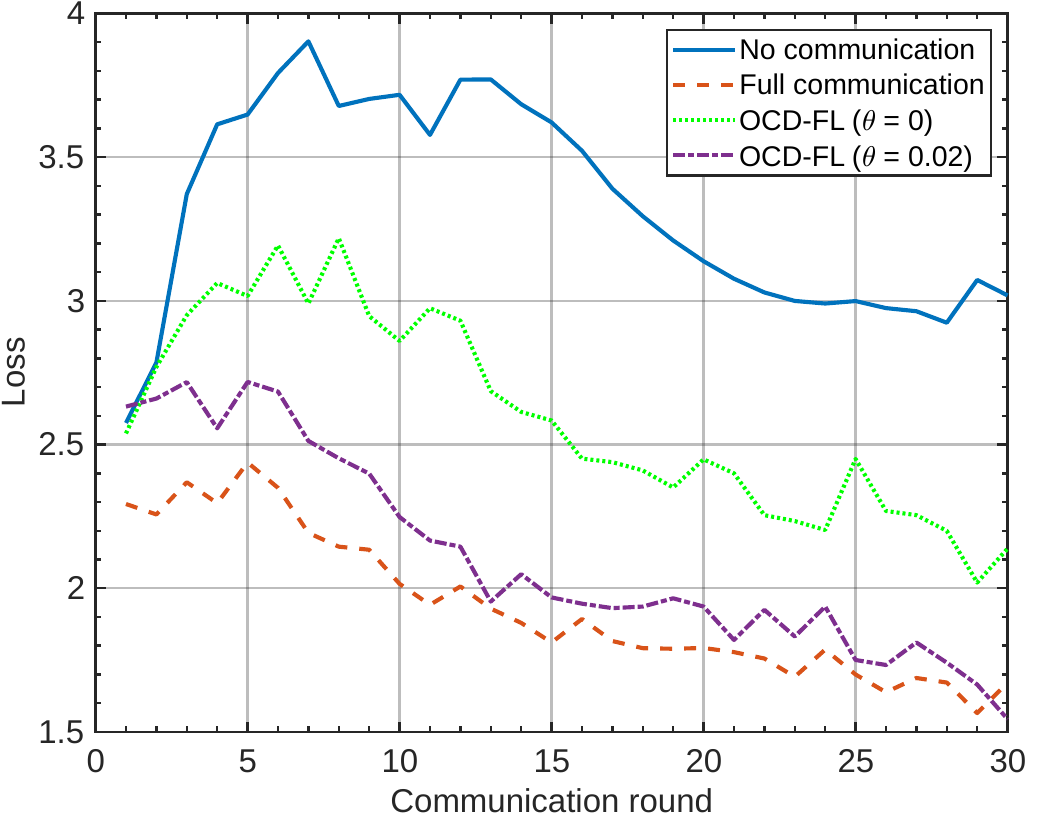}\label{fig:cifar-niid-loss}}%
            
            \caption{Avg. accuracy and loss (CIFAR-10, different schemes).}
            \label{fig:cifar-eval}
        \end{figure}

\vspace{-10pt}    
    \subsection{Simulation Results}

OCD-FL is evaluated against baseline schemes ``No communication'' where each client trains exclusively locally, and ``Full communication'' where any node communicates with all its neighbors. \textcolor{blue}{Note that we considered ``OCD-FL ($\theta=0$)'' since it is analogous to the lower case of (\ref{eq:main-opt}) where $\sum_{k\in\mathcal{K}_i} \sigma(w_k)=1$, while ``Full communication'' reflects the upper case of (\ref{eq:main-opt}) where $\sum_{k\in \mathcal{K}_i} \sigma(w_k)=K_i$. 
}


In Fig. \ref{fig:mnist-eval}, we illustrate the FL performances, in terms of accuracy and loss, for the proposed method when applied to the MNIST dataset, and compared to the benchmarks, under IID and non-IID scenarios. For the IID scenario, the proposed OCD-FL method ($\theta=0$ and $\theta=0.02$) outperforms all benchmarks. Indeed, by setting $\theta$ to 0.02, we introduce a regularization term that promotes collaboration with a wider range of neighbors. Such results highlight the importance of controlled collaboration between nodes to achieve consensus on efficient models. In the meanwhile, the ``No communication'' scheme provides the worst results. This is expected since each client relies only on its knowledge for training. In the non-IID scenario, the performance gap between the ``No collaboration'' scheme and the other ones increases. This is due to the higher complexity of training in the non-IID setting. Moreover, the proposed OCD-FL method ($\theta=0.02$) still outperforms all other methods, while the performance of OCD-FL ($\theta=0$) degrades below that of ``Full communication''. Indeed, since $\theta=0$, regularization is eliminated. Hence, our scheme limits its peer selection for each node to a small number, which may not be sufficient to train efficiently in the non-IID scenario. Indeed, our scheme struggles to achieve a consensus on an efficient model, and the instability of the associated learning curve highlights the network's inability to generalize. \textcolor{blue}{Note that the initial increasing trend in the loss curve is due to gradient instability during the first few rounds. The increase is less significant under ``Full Communication" and ``OCD-FL $(\theta=0.02)$" since efficient model aggregation contributes to faster gradient stability. This increase is not observed on the MNIST simpler dataset, where it only takes the gradients on a small number of training rounds to stabilize.}

        Fig. \ref{fig:cifar-eval} presents the same results as in Fig. \ref{fig:mnist-eval}, but for the CIFAR-10 dataset. As it can be seen, for the IID scenario, ``OCD-FL ($\theta=0.02$)'' is capable of providing similar performances, in terms of accuracy and loss, to ``Full communication'', while the gap with ``OCD-FL ($\theta=0$)'' and ``No communication'' is very significant. For instance, after 20 rounds, the gap in accuracy is approximately 10\%. In the non-IID scenario, ``Full communication'' presents the best performances, while ``OCD-FL ($\theta=0.02$)'' falls slightly behind, by about 2\% in terms of accuracy. 
        ``OCD-FL ($\theta=0$)'', although not the best scheme, is still significantly outperforming ``No communication''. Note that, even though our scheme is not the best in CIFAR-10 with non-IID, an optimal $\theta$ might be determined, which would provide very close performances to the ``Full communication'' scheme.   

\begin{figure}[t]
            \centering
            \subfloat[MNIST]{\includegraphics[width=0.5\linewidth]{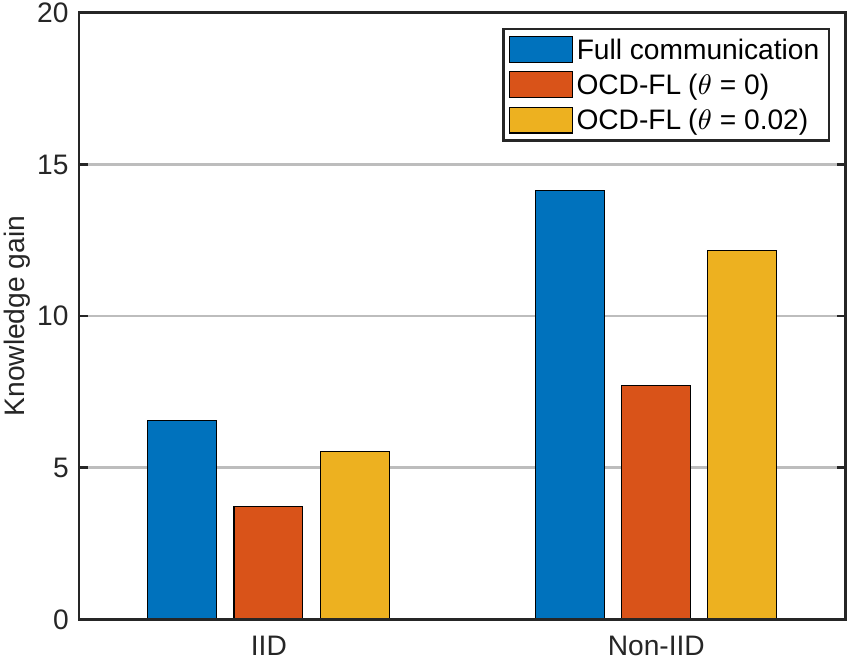}\label{fig:kg-mnist}}%
            \hfil
            \subfloat[CIFAR-10]{\includegraphics[width=0.5\linewidth]{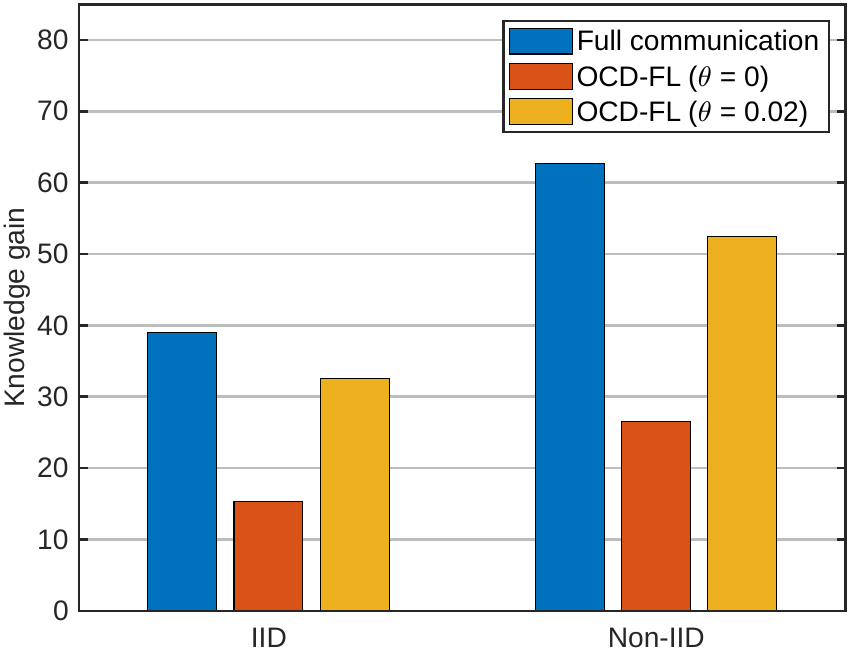}\label{fig:kg-cifar}}%
            \caption{\textcolor{blue}{Knowledge gain (different schemes and scenarios).}}
            \label{fig:kg-eval}
        \end{figure}

    \textcolor{blue}{Fig. \ref{fig:kg-eval} presents the average knowledge gain under different scenarios. Although OCD-FL $(\theta = 0)$ fails to promote knowledge sharing between clients, it manages to compete with ``Full Communication'' when applied with regularization $\theta=0.02$, in IID and non-IID settings. With non-IIDness, the resulting knowledge gain is significantly high. This is due to the model performance disparity between clients.}

        \begin{figure}[t]
            \centering
            \subfloat[MNIST]{\includegraphics[width=0.5\linewidth]{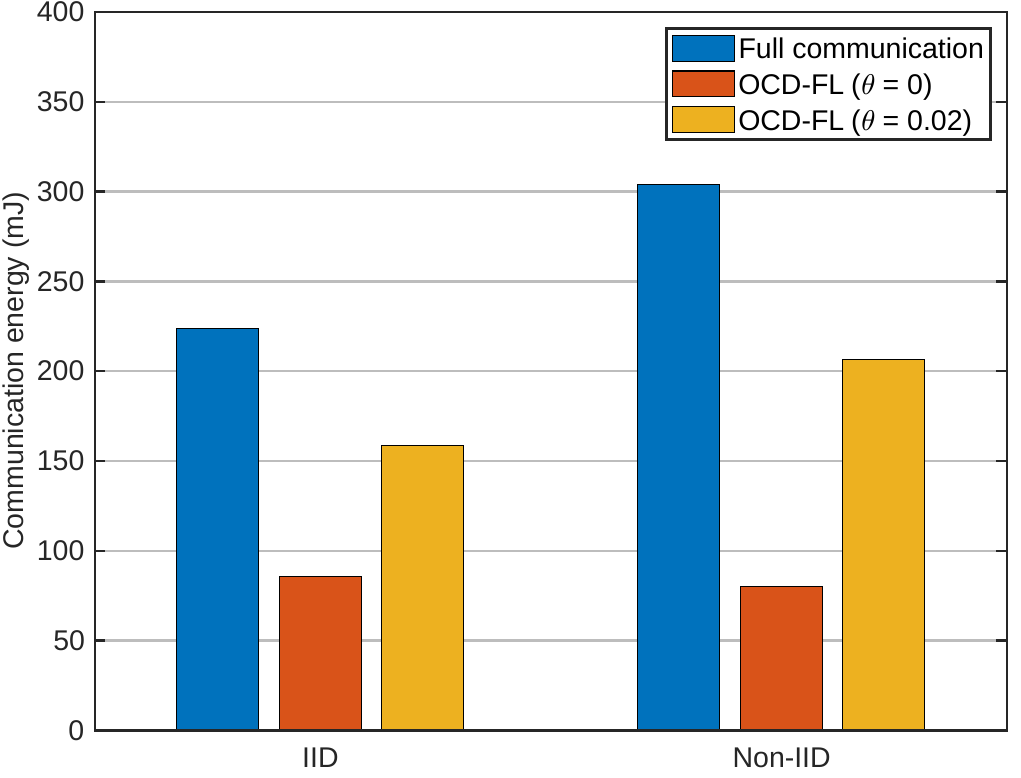}\label{fig:energy-mnist}}%
            \hfil
            \subfloat[CIFAR-10]{\includegraphics[width=0.5\linewidth]{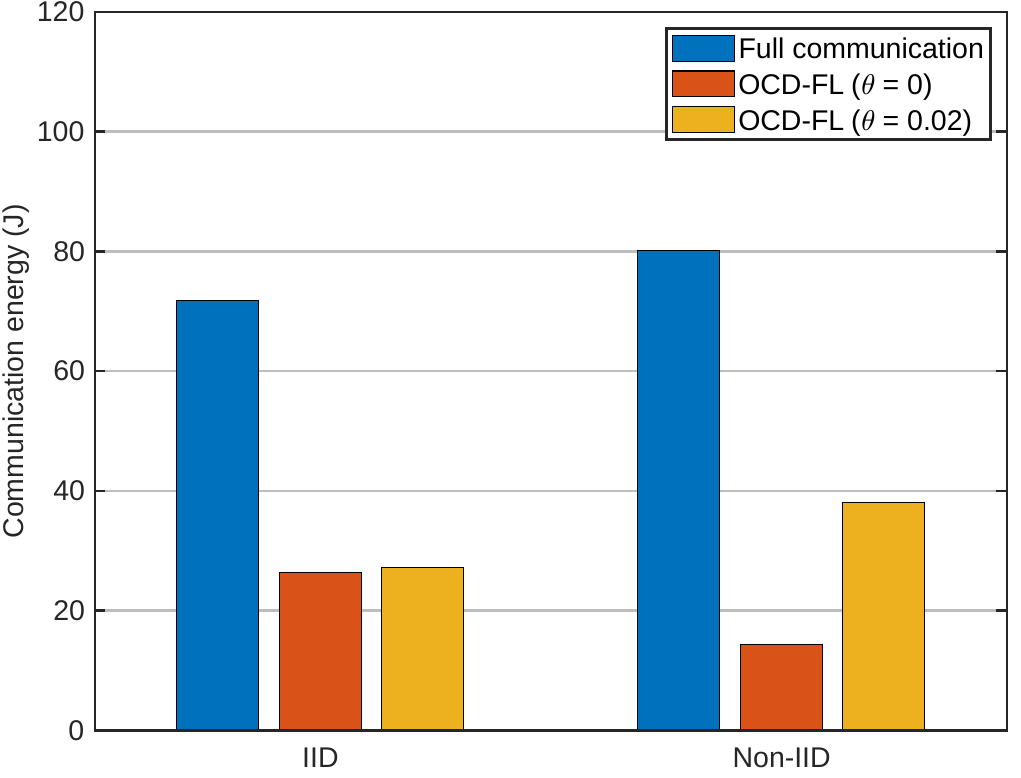}\label{fig:energy-cifar}}%
            \caption{Consumed communication energy (different schemes and scenarios).}
            \label{fig:energy-eval}
        \end{figure}
        
        \textcolor{blue}{Similarly,} in Fig. \ref{fig:energy-eval}, we depict the communication energy consumed by each system with OCD-FL or ``Full communication'' in IID and non-IID scenarios, and for MNIST and CIFAR-10 datasets. 
        ``Full communication'' consumed the highest amounts of energy in any setting, since it relies on communications between all $N$ clients. In contrast, our OCD-FL scheme consumes less energy between 30\% and 80\% than ``Full communication''. This is mainly due to the accurate selection of peers for model sharing.

        

\vspace{-10pt}
\section{Conclusion} \label{sec:conclusion}
    In this paper, we proposed a novel distributed FL scheme, called OCD-FL. The latter systematically selects neighbors for peer-to-peer FL collaboration. Our solution incorporates a trade-off between knowledge gain and energy efficiency. To do so, the developed peer selection strategy was assimilated into a regularized multi-objective optimization problem aiming to maximize knowledge gain while consuming minimum energy. The OCD-FL method was evaluated in terms of FL accuracy, loss, and energy consumption, and compared against baselines and under several scenarios. OCD-FL proved its capability to achieve consensus on an efficient FL model while significantly reducing communication energy consumption between 30\% and 80\%, compared to the best benchmark. \textcolor{blue}{Although we conducted a comprehensive evaluation of OCD-FL, several aspects of the network's layout present interesting research opportunities, such as the impact of a time-varying topology on model convergence.} \textcolor{blue}{Moreover, despite the adoption of federated averaging in our work, this research serves as a proof of concept and lays the groundwork for future exploration of other distributed FL systems, where different FL aggregation techniques might be experimented.
}



\ifCLASSOPTIONcaptionsoff
  \newpage
\fi
\vspace{-10pt}
\bibliographystyle{IEEEtran}  
\bibliography{references}



\end{document}